\documentclass[a4paper, 10pt, conference]{ieeeconf}      
\usepackage{FG2024}
\usepackage{url}
\usepackage{color}
\usepackage{multirow}
\usepackage{booktabs}
\usepackage{amsmath}
\usepackage{graphicx}

\FGfinalcopy 

\IEEEoverridecommandlockouts                              
\def\FGPaperID{****} 

\title{\LARGE \bf
REACT 2024: the Second Multiple Appropriate Facial Reaction Generation Challenge
}

\author{\parbox{16cm}{\centering
    {\large Siyang Song$^{1,3*\dag}$, Micol Spitale$^{2,3*}$, Cheng Luo$^{4*}$, Cristina Palmero$^{5}$, German Barquero$^{5}$, Hengde Zhu$^{1}$, Sergio Escalera$^{5}$, Michel Valstar$^{6}$, Tobias Baur$^{7}$, Fabien Ringeval$^{8}$, \\ Elisabeth André$^{7}$, and Hatice Gunes$^{1}$ }\\
    {\normalsize
    $^1$ University of Leicester, UK; $^2$ Politecnico di Milano, Italy \\
    $^3$ University of Cambridge, UK; $^4$ Monash University, Australia \\
    $^5$ Universitat de Barcelona, Spain; $^6$ University of Nottingham, UK  \\
    $^7$ University of Augsburg, Germany; $^8$ Université Grenoble Alpes, France}}
    \thanks{$*$ Equal Contribution; $\dag$ Corresponding Author}
}

\begin{document}

\ifFGfinal
\thispagestyle{empty}
\pagestyle{empty}
\else
\author{Anonymous FG2024 submission\\ Paper ID \FGPaperID \\}
\pagestyle{plain}
\fi
\maketitle

\begin{abstract}

In dyadic interactions, humans communicate their intentions and state of mind using verbal and non-verbal cues, where multiple different facial reactions might be \textit{appropriate} in response to a specific speaker behaviour. Then, how to develop a machine learning (ML) model that can automatically generate multiple appropriate, diverse, realistic and synchronised human facial reactions from an previously unseen speaker behaviour is a challenging task. Following the successful organisation of the first REACT challenge (REACT 2023), this edition of the challenge (REACT 2024) employs a subset used by the previous challenge, which contains segmented 30-secs dyadic interaction clips originally recorded as part of the NOXI and RECOLA datasets, encouraging participants to develop and benchmark Machine Learning (ML) models that can generate multiple \textit{appropriate} facial reactions (including facial image sequences and their attributes) given an input conversational partner's stimulus under various dyadic video conference scenarios. This paper presents: (i) the guidelines of the REACT 2024 challenge; (ii) the dataset utilized in the challenge; and (iii) the performance of the baseline systems on the two proposed sub-challenges: Offline Multiple Appropriate Facial Reaction Generation and Online Multiple Appropriate Facial Reaction Generation, respectively. The challenge baseline code is publicly available at \url{https://github.com/reactmultimodalchallenge/baseline_react2024}. 

\end{abstract}

\section{INTRODUCTION}


Recent years have seen an increasing number of studies targeting human-human dyadic interaction analysis~\cite{barquero2022didn}. Previous studies \cite{huang2017dyadgan,song2019exploiting,yoon2022genea} have investigated the problem of automatically generating a specific reaction that resembles the ground-truth (real) response or reaction for a given input. Most of these studies proposed deterministic approach that aims to reproduce the ground-truth reaction without considering the non-verbal aspects that enrich the message conveyed. Few studies have looked into the generation of appropriate reactions as non-verbal behaviours, with a main focus on generating a \textit{single appropriate} reaction, e.g., hand gesture \cite{yoon2022genea}, facial reaction \cite{huang2017dyadgan,song2022learning,shao2021personality,ng2022learning,zhou2022responsive}, or full-body postures~\cite{barquero22comparison}.

As discussed in \cite{song2023multiple}, given a human behaviour (called speaker behaviour), multiple appropriate facial reactions could be expressed by not only different individuals but also the same individual under different situations in response to it. Consequently, a Multiple Appropriate Facial Reaction Generation (MAFRG) task has been proposed. The REACT 2024 Challenge is the second competition event aimed at comparison of machine learning methods for MAFRG tasks, with all participants competing under strictly the same conditions. The REACT 2024 Challenge follows the similar purpose and form as the REACT 2023 challenge \cite{song2023react2023}, focusing on two MAFRG tasks: offline and online Multiple Appropriate Facial Reaction Generation (offline and online MAFRG).


Although the organization of the REACT 2023 challenge facilitated the creation of several successful solutions \cite{xu2023reversible,luo2023reactface,yu2023leveraging,liang2023unifarn,hoque2023beamer,xu2023mrecgen,barquero2022belfusion} for both online and offline MAFRG tasks, most of them were not able to provide realistically generated facial reaction sequences but only focussed on generating facial attributes of the predicted facial reactions. Hence, this edition aims to promote the submission of results that include realistic facial reaction video clips. To assist in addressing the  challenge of generating facial reaction video clips, this edition focuses specifically on video-conference settings and therefore includes only the NoXI \cite{cafaro2017noxi} and RECOLA \cite{ringeval2013introducing} datasets, due to the more noisy data of in-person settings (i.e., a major reason for excluding the UDIVA dataset \cite{palmero2021context} in this edition that was used in REACT 2023 challenge).

The REACT 2024 Challenge adopts the metrics defined in \cite{song2023multiple} to evaluate four aspects of the submitted models in terms of their generated facial reactions, namely: appropriateness, diversity, realism and synchrony. Participants are required to submit their developed model, checkpoints and well-explained source code, accompanied by a paper submitted to the REACT 2024 Challenge describing their proposed methodology and the achieved results. Only contributions that meet the pre-determined requirements, terms and conditions \footnote{https://sites.google.com/cam.ac.uk/react2024/home} are eligible for participation. The organisers do not engage in active participation themselves, but instead undertake a re-evaluation of the findings of the systems submitted to both sub-challenges. Differently from the previous edition, the ranking of the submitted models in this challenge depend on two metrics: Appropriate facial reaction correlation (FRCorr) of the generated facial reaction attributes and facial reaction realism (FRRea) of the generated facial reaction video clips, for both sub-challenges. In addition, participants should also report Facial reaction distance (FRDist), facial reaction diverseness (FRDiv), Facial reaction variance (FRVar), Diversity among facial reactions generated from different speaker behaviours (FRDvs) and Synchrony between generated facial reactions and speaker behaviours (FRSyn).

\section{Challenge Corpora}

The REACT 2024 challenge employs two video conference corpora: NoXi \cite{cafaro2017noxi} and RECOLA \cite{ringeval2013introducing}. Specifically, we first segmented each audio-video clip in two datasets into a 30-seconds long clip as in \cite{ambady1992thin,song2023react2023}. Then, we cleaned the dataset by selecting only the dyadic interactions with complete data of both conversational partners (where both faces were within the frame of the camera). This resulted in 5919 clips of 30 seconds each (71,8 hours of audio-video clips), specifically: 5870 clips (49 hours) from the NoXi dataset and 54 clips (0,4 hour) from the RECOLA dataset. We divided the datasets into training (1,585 video clips from NoXI and 9 video clips from RECOLA), test (797 video clips from NOXI and 9 video clips from RECOLA) and validation (553 from NOXI and 9 from RECOLA) sets. We split the datasets with a subject-independent strategy (i.e., the same subject was never included in the training/validation and test sets). In this challenge, 25 frame-level facial attributes are provided for each facial frame, namely 15 AUs' occurrence (AU1, AU2, AU4, AU6, AU7, AU9, AU10, AU12, AU14, AU15, AU17, AU23, AU24, AU25 and AU26) predicted by the state-of-the-art GraphAU model \cite{luo2022learning,song2022gratis}, as well as 2 facial affects (i.e., valence and arousal intensities) and 8 facial expression probabilities (i.e., Neutral, Happy, Sad, Surprise, Fear, Disgust, Anger and Contempt) predicted by \cite{toisoul2021estimation}.

\section{Evaluation Metrics}

In this challenge, the submitted models are expected to generate two types of outputs for representing each facial reaction: (i) 25 facial attribute time-series; and (ii) a 2D facial image sequence. We followed \cite{song2023multiple,song2023react2023} to comprehensively evaluate three aspects of the generated facial reaction attributes: (i) \textbf{Appropriateness} based on two metrics, \textbf{FRCorr}: Concordance Correlation Coefficient (CCC) and \textbf{FRDist}: Dynamic Time Warping (DTW); (ii) \textbf{Diversity}: \textbf{FRVar}, \textbf{FRDiv}, and \textbf{FRDvs}; and (iii) \textbf{Synchrony}: the Time Lagged Cross Correlation (TLCC), called \textbf{FRSyn} in this challenge. Also, the \textbf{Realism} of the generated facial reaction video clips is assessed using the Fréchet Inception Distance (FID), denoted as \textbf{FRRea}.

\section{Baseline systems}

\begin{figure}
    \centering
    \includegraphics[width=\linewidth]{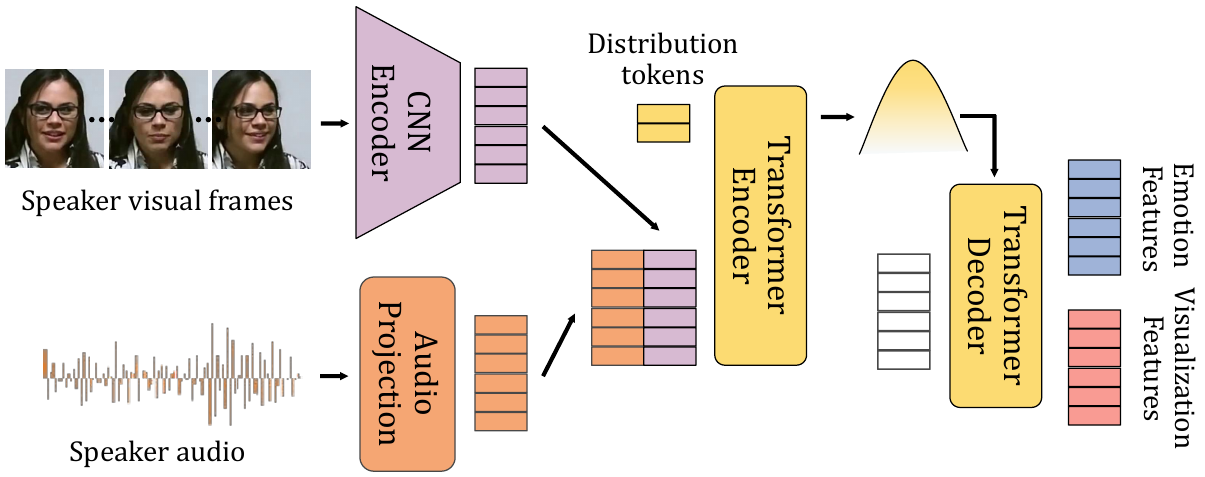}
    \caption{Overview of the Trans-VAE baseline.}
    \label{fig:baseline_Trans-VAE}
\end{figure}

\begin{figure}
    \centering
    \includegraphics[width=\linewidth]{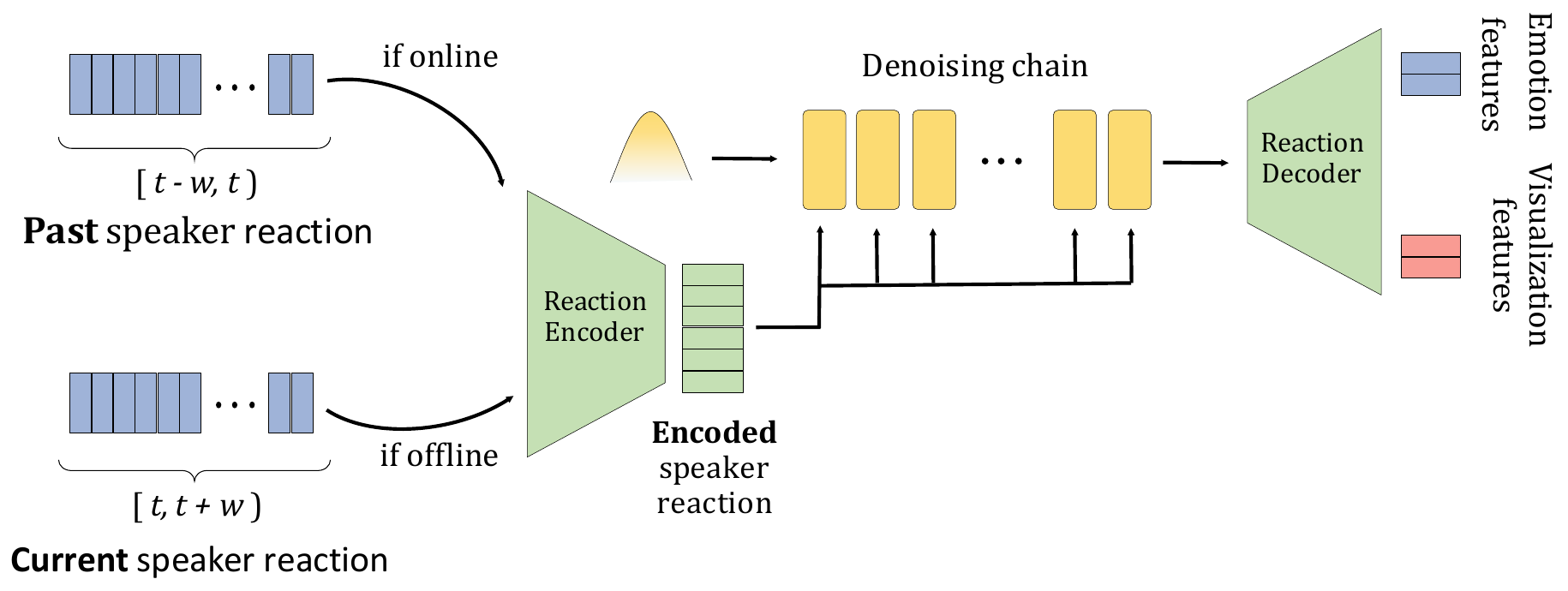}
    \caption{Overview of the BeLFusion baseline.}
    \label{fig:baseline_belfusion}
\end{figure}

\begin{figure}
    \centering
    \includegraphics[width=\linewidth]{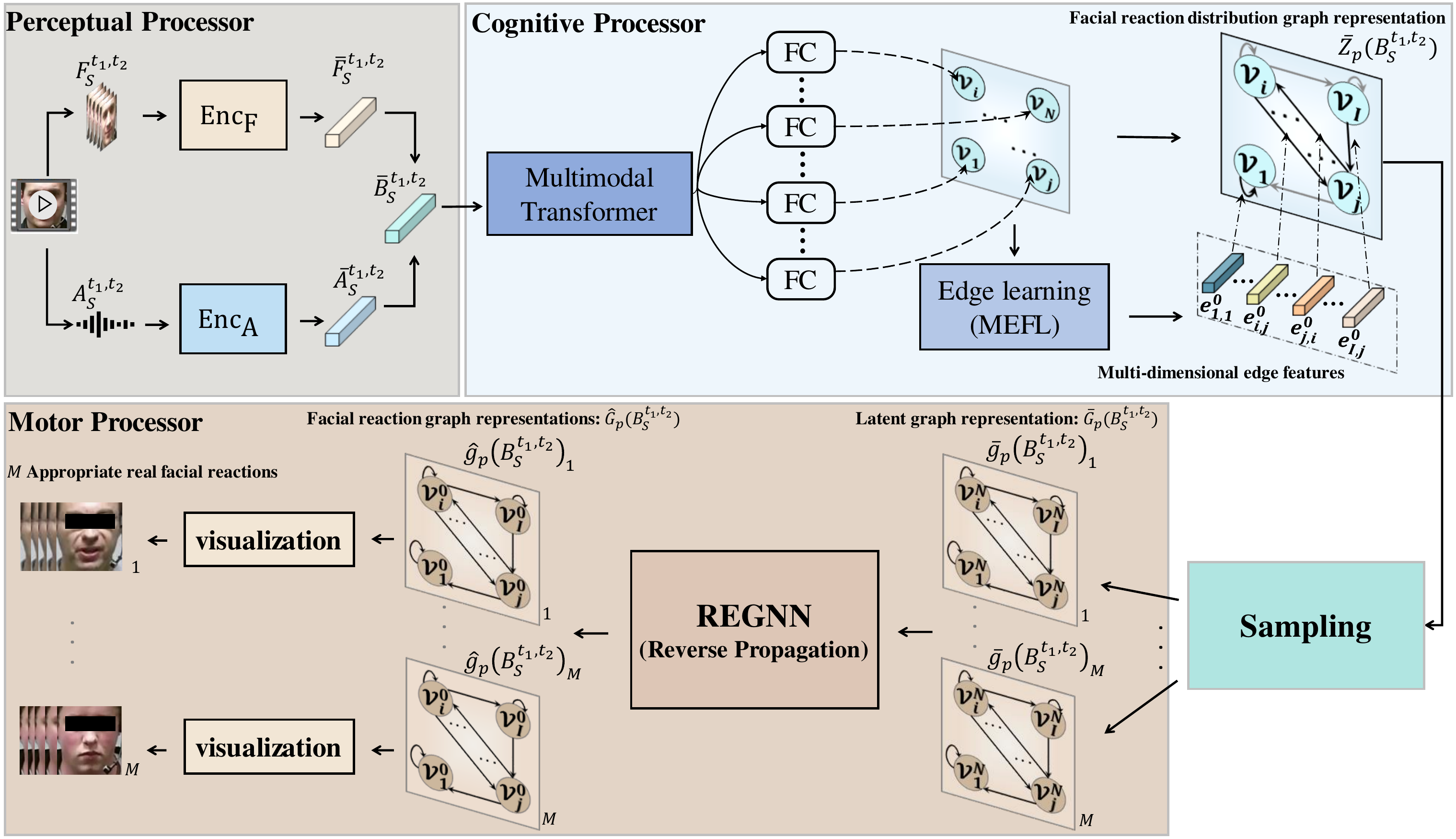}
    \caption{Overview of the REGNN baseline \cite{xu2023reversible}.}
    \label{fig:baseline_REGNN}
\end{figure}

\textbf{Trans-VAE:} We re-employ the same Trans-VAE baseline used in previous challenge \cite{song2023react2023} to this challenge. This baseline is inspired by \cite{luo2023reactface}, which follows the similar architecture as the TEACH \cite{athanasiou2022teach}. As shown in Fig.~\ref{fig:baseline_Trans-VAE}, it is made up of (i) a \textbf{CNN encoder} that extract facial reaction-related features from the input speaker facial image sequence; (ii) a \textbf{transformer encoder} that combines the learned facial embeddings and baseline audio embeddings (78-dimnesional MFCC features) extracted from the speaker audio behaviours using Torchaudio library \cite{yang2022torchaudio}, based on which an Gaussian Distribution is predicted to describe multiple appropriate facial reactions of the input speaker behaviour; and (iii) a \textbf{transformer decoder} that samples two types of facial reaction representations from the predicted distribution: 1) a set of 3D Morphable Model (3DMM) coefficients (i.e., 52 facial expression coefficients, 3 pose coefficients and 3 translation coefficients defined by \cite{wang2022faceverse}) and 2) an multi-channel facial attribute time-series (i.e., 25-channel time-series including 15 frame-level AUs' occurrence, 8 frame-level facial expression probabilities as well as frame-level valence and arousal intensities). Please refer to \cite{song2023react2023} for the detailed description of applying this baseline for online and offline MAFRG tasks.

\textbf{BeLFusion.} We also re-use BeLFusion without behavioural disentanglement as our second baseline~\cite{barquero2022belfusion}, see Fig.~\ref{fig:baseline_belfusion}. Its training involves two steps. First, a variational autoencoder (VAE) learns a lower representation of the sequence of visual features (e.g., AUs, facial affects, and expressions) for $w$ frames. A regressor is incorporated after the VAE's decoder to transform the decoded reaction into a sequence of 3DMM coefficients. Second, a latent diffusion model (LDM) is optimized to, when given the speaker's reaction, predict the lower-dimensional representation of the listener's appropriate facial reaction. BeLFusion employs a window-based approach where the $T/w$ reactions are predicted independently. Afterwards, the $w$-frames-long $T/w$ reactions are arranged to construct the complete reaction. As in \cite{song2023react2023}, the listener's visual features for the window $[t, t+w)$ is conditioned on the past speaker's features at $[t-w, t)$. Features for the segment $[0, w)$ are all set to zeroes. In the offline subchallenge, the generation is conditioned on the speaker's features within the same time period: $[t, t+w)$. The LDM's loss is the average of the latent and reconstructed MSE losses, and the denoising chain length is set to 10 steps.

\textbf{Reversible Graph Neural Network (REGNN):} We also employ the REGNN-based MAFRG approach \cite{xu2023reversible} as the second baseline. As illustrated in Fig.~\ref{fig:baseline_REGNN}, it consists of three main modules: (i) a \textbf{Perceptual Processor} that encodes the input speaker audio-facial behaviour as a pair of latent audio and facial representations; (ii) a \textbf{Cognitive Processor ($\text{Cog}$}) that predicts a Gaussian Mixture Graph Distribution describing all appropriate facial reactions in response to the input speaker behaviour; and (iii) an Reversible GNN-based \textbf{Motor Processor} that samples an appropriate facial reaction from the predicted appropriate facial reaction distribution. During the training, the Reversible GNN employed in this approach encodes all appropriate facial reactions of each input speaker behaviour as an ground-truth appropriate facial reaction distribution, enforcing the cognitive processor to predict the same distribution from the speaker representations obtained by the perceptual processor. As a result, the \textit{one-to-many mapping} training problem is re-formulated as a \textit{one-to-one mapping} problem. Please refer to \cite{xu2023reversible} for more implementation details of its offline MAFRG system.

\section{Baseline Results}

It is clear that all three baselines outperformed the B\_Random, B\_Mime, B\_MeanSeq and B\_MeanFr, suggesting that they can predict meaningful appropriate human facial reactions from different speaker behaviours despite that the predicted facial reactions performances are not very solid.  

\begin{figure*}[htb!]
    \centering
    \includegraphics[width=1.5\columnwidth]{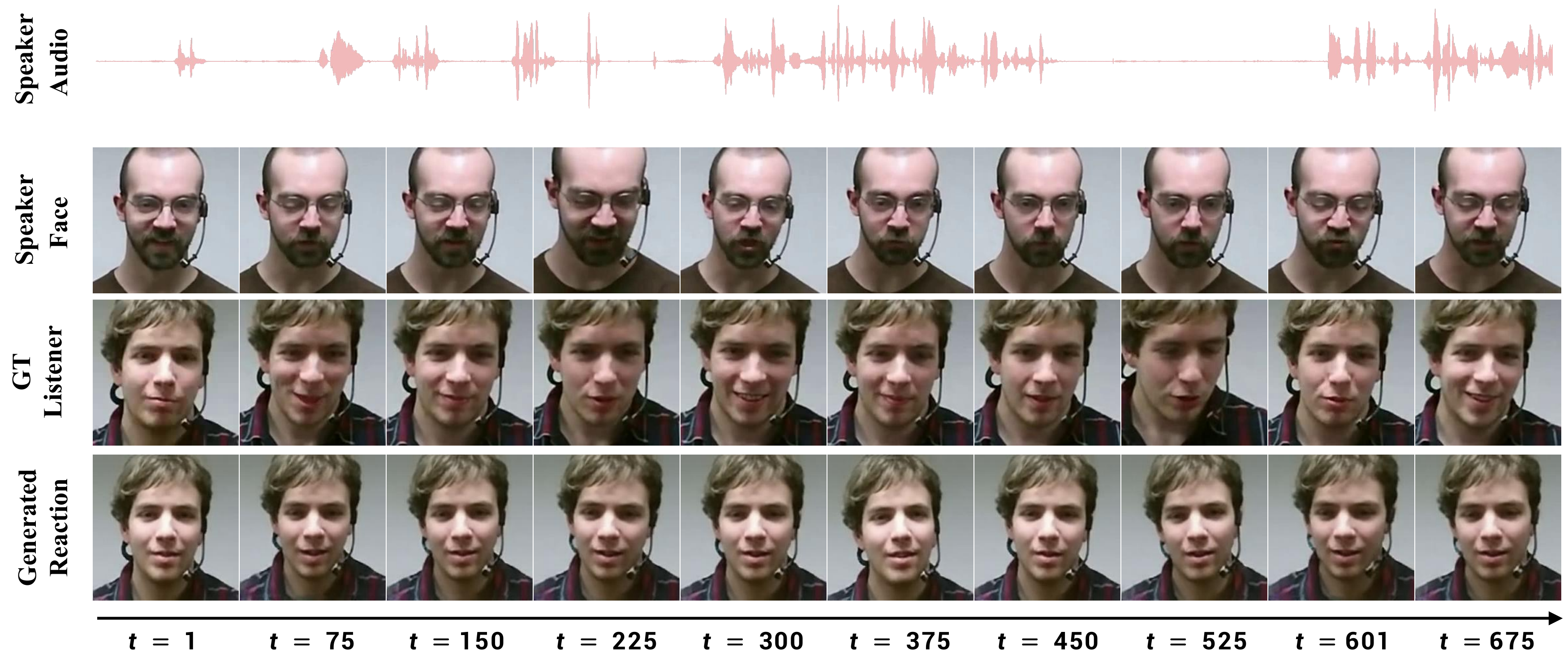} 
        \caption{Examples of generated  listener reactions to a given speaker behaviour (including the speaker's audio and face frames). These reactions are generated by an offline Trans-VAE model.  }
    \label{fig:ex_vis_offline}
\end{figure*}

\textbf{Trans-VAE baseline:} The Trans-VAE baseline model serves as a fundamental baseline for comparison purposes. This baseline demonstrates the capability to generate facial reactions that exhibit a modest level of diversity, as measured by metrics such as FRDiv, FRVar, and FRDvs, alongside a moderate level of appropriateness and comparable synchronization in both offline and online scenarios. In contrast to random facial reactions (B\_Random), the Trans-VAE model achieved higher appropriateness (FRD). Furthermore, it surpasses in generating more diverse samples (FRDiv) compared to replicating facial sequences mirroring the speaker's facial behaviour. We visualise example facial reactions generated by this baseline in Fig. \ref{fig:ex_vis_offline}.

\textbf{BeLFusion baseline:} While BeLFusion shows a performance similar to Trans-VAE in terms of accuracy (FRD), it generates more diverse reactions. The competitive performance of such baseline without access to raw audio or video data highlights the need for better multimodal approaches tailored for this application. We also observe that binarizing the action units predicted greatly improves the diversity, but penalizes the accuracy and synchrony. The similar results in both online and offline scenarios suggest that a window-based approach might be insufficient to exploit all the information available in the visual features.

\textbf{Reversible Graph Neural Network (REGNN) baseline:} In the offline task evaluation, REGNN demonstrates clear advantages over Trans-VAE and BeLFusion baselines in terms of the appropriateness metrics, as indicated by the highest FRCorr and lowest FRD. In addition, the facial reactions generated by REGNN are more synchronised with the speaker behaviour. When it comes to the diversity, REGNN is less effective in generating diverse facial reactions, compared to the BeLFusion baseline.

\begin{table*}[htb!]\centering
 \footnotesize
 \caption{\label{tb:test_quantitative_offline} 
Baseline offline and online facial reaction generation results achieved on the test set.} 
\begin{tabular}{lccccccc}
\toprule 

    \multirow{2}{*}{\textbf{Method}} & \multicolumn{2}{c}{\textbf{Appropriateness}}  & \multicolumn{3}{c}{\textbf{Diversity}} &\textbf{Realism}  &   \textbf{Synchrony}\\
    \cmidrule(r){2-3} \cmidrule(r){4-6}  \cmidrule(r){7-7}  \cmidrule(r){8-8} 
    & \textbf{FRCorr} ($\uparrow$) &  \textbf{FRDist} ($\downarrow$) &  \textbf{FRDiv} ($\uparrow$) &  \textbf{FRVar} ($\uparrow$) &  \textbf{FRDvs} ($\uparrow$)  & \textbf{FRRea} ($\downarrow$)    & \textbf{FRSyn} ($\downarrow$)\\ 
    \midrule
    GT & 8.73 & 0.00 & 0.0000 & 0.0724 & 0.2483 & 53.96 & 47.69 \\ \midrule
    
    B\_Random & 0.05 & 237.21 & 0.1667 & 0.0833 & 0.1667 & - & 43.84 \\

    B\_Mime & 0.38 & 92.94 & 0.0000 & 0.0724 & 0.2483 & - & 38.54 \\
    B\_MeanSeq & 0.01 & 97.13 & 0.0000 & 0.0000 & 0.0000 & - & 45.28 \\
    B\_MeanFr & 0.00 & 97.86 & 0.0000 & 0.0000 & 0.0000 & - & 49.00 \\
    \midrule
    \textbf{Offline Results}\\
    \midrule
    Trans-VAE   & 0.03 &92.81 & 0.0008 & 0.0002 & 0.0006 & 67.74 & 43.75\\
        
    BeLFusion ($k$=1)& 0.10 & 92.32 & 0.0068 & 0.0073 & 0.0094 & - & 44.94 \\
    BeLFusion ($k$=10) & 0.12 & 91.60 & 0.0105 & 0.0082 & 0.0116 & - & 44.87 \\
    BeLFusion ($k$=10) + Binarized AUs & 0.12 & 94.16 & 0.0360 & 0.0249 & 0.0384 & - & 49.00 \\
    REGNN   & 0.19 & 84.54 & 0.0007 & 0.0061 & 0.0342 & - & 41.35 \\
    \midrule
\textbf{Online Results}\\
\midrule
  
    Trans-VAE   & 0.07 & 90.31 & 0.0064 & 0.0012 & 0.0009 & 69.19 & 44.65\\
    BeLFusion ($k$=1) & 0.12 & 91.11 & 0.0083 & 0.0079 & 0.0103 & - & 45.17 \\
    BeLFusion ($k$=10)  & 0.12 & 91.45 & 0.0112 & 0.0082 & 0.0120 & - & 44.89 \\
    BeLFusion ($k$=10) + Binarized AUs & 0.12 & 94.09 & 0.0379 & 0.0248 & 0.0397 & - & 49.00 \\
\bottomrule
\end{tabular}
\label{tb:test_results}
\end{table*}

\section{Participation and Conclusion}

This paper introduced REACT 2024 Challenge in conjunction with the IEEE International Conference on Automatic Face and Gesture Recognition (FG) 2024, which focuses on multiple appropriate facial reaction generation under various video conference-based dyadic interactions scenarios. 
A total of 13 teams from 6 countries registered for this challenge, with 12 teams participating in the offline MAFRG sub-challenge and 13 teams participating in the online MAFRG sub-challenge. 
Our evaluation protocol strictly will rank all participant models under the same settings by comprehensively considering two aspects of their generated facial reactions: appropriateness, diversity, realism and synchrony. 
We hope that both the challenge data and code, as well as the systems and results of the competing teams, will serve as a valuable stepping stone for researchers and  practitioners interested in the area of generative AI and automatic facial reaction generation.
Our future efforts will be directed at continuing to organize REACT challenges in conjunction with well-known conferences while introducing new datasets and new modalities.

\section*{Acknowledgements}
H. Gunes is supported by the EPSRC/UKRI under grant ref. EP/R030782/1 (ARoEQ). M. Spitale is supported by PNRR-PE-AI FAIR project funded by the NextGeneration EU program.

{\small
\bibliographystyle{ieee}
\bibliography{egbib}
}

\end{document}